\documentclass[sigconf]{acmart}
\usepackage{subcaption}
\usepackage{algorithm}
\usepackage{algorithmic}
\usepackage{multirow}
\usepackage{makecell}
\AtBeginDocument{%
  }

\setcopyright{acmlicensed}
\copyrightyear{2025}
\acmYear{2025}
\acmDOI{XXXXXXX.XXXXXXX}
\acmConference[CIKM '25]{Make sure to enter the correct
  conference title from your rights confirmation email}{Nov 10--14,
  2025}{Seoul, KR}
\acmISBN{978-1-4503-XXXX-X/2018/06}

\begin{document}

\title{Caption, Create, Continue: Continual Learning with Pre-trained Generative Vision-Language Models}

\author{Indu Solomon}
\affiliation{ 
\institution{International Institute of Information Technology Bangalore (IIITB), India}
\city{} 
\country{}
}\email{indu.solomon@iiitb.ac.in}

\author{Aye Phyu Phyu Aung}
\affiliation{
\institution{Institute for Infocomm Research, Agency for Science, Technology and Research (A*STAR), Singapore}
\city{}
\country{}
}
\email{aye_phyu_phyu_aung@i2r.a-star.edu.sg}

\author{Uttam Kumar}
\affiliation{
\institution{International Institute of Information Technology Bangalore (IIITB), India}
\city{}
\country{}
}
\email{uttam@iiitb.ac.in}
\author{Senthilnath Jayavelu}
\affiliation{
\institution{Institute for Infocomm Research, Agency for Science, Technology and Research (A*STAR), Singapore}
\city{}
\country{}
}
\email{J_Senthilnath@i2r.a-star.edu.sg}

\renewcommand{\shortauthors}{Trovato et al.}

\begin{abstract}
Continual learning (CL) enables models to adapt to evolving data streams without catastrophic forgetting, a fundamental requirement for real-world AI systems. However, the current methods often depend on large replay buffers or heavily annotated datasets which are impractical due to storage, privacy, and cost constraints.
We propose CLTS (\textbf{C}ontinual \textbf{L}earning via \textbf{T}ext-Image \textbf{S}ynergy), a novel class-incremental framework that mitigates forgetting without storing real task data. CLTS leverages pre-trained vision-language models, BLIP (Bootstrapping Language-Image Pre-training) for caption generation and stable diffusion for sample generation. Each task is handled by a dedicated Task Head, while a Task Router learns to assign inputs to the correct Task Head using the generated data.
On three benchmark datasets, CLTS improves average task accuracy by up to 54$\%$ and achieves 63 times better memory efficiency compared to four recent continual learning baselines, demonstrating improved retention and adaptability. CLTS introduces a novel perspective by integrating generative text-image augmentation for scalable continual learning.
\end{abstract}

% <concept_id>10010405.10010432.10010441</concept_id>
% <concept_desc>Applied computing~Physics</concept_desc>

% \ccsdesc[500]{Applied computing~Physics}
\begin{CCSXML}
<ccs2012>
 <concept>
  <concept_id>10010147.10010178.10010179</concept_id>
  <concept_desc>Computing methodologies~Machine learning approaches</concept_desc>
  <concept_significance>500</concept_significance>
 </concept>
 <concept>
  <concept_id>10010147.10010178.10010179.10010182</concept_id>
  <concept_desc>Computing methodologies~Continual learning</concept_desc>
  <concept_significance>300</concept_significance>
 </concept>
</ccs2012>
\end{CCSXML}

\ccsdesc[500]{Computing methodologies~Machine learning approaches}
\ccsdesc[300]{Computing methodologies~Continual learning}

\keywords{Pre-trained model, Task, Class-incremental, Catastrophic forgetting, Continual learning.}

\maketitle

\section{Introduction}
Conventional machine learning (ML) models are typically trained on curated datasets and often fail when exposed to evolving data distributions. This leads to mis-classification, as the model forgets previously learnt knowledge when adapting to new data, a phenomenon known as catastrophic forgetting \cite{parisi2019continual, pratama2021unsupervised}. While re-training on the entire dataset can mitigate this, but it is often impractical due to computational, storage and privacy constraints. Continual learning (CL) addresses this issue by enabling models to incrementally learn from data streams without forgetting prior knowledge \cite{de2021continual}. A stream of data arriving at a specific time is referred to as a task and CL settings are broadly classified into Task Incremental Learning (Task-IL), Domain Incremental Learning (Domain-IL), and Class Incremental Learning (Class-IL) \cite{zhou2024class, van2022three}. Task-IL assumes the task identity is known during inference, while Domain-IL introduces distribution shifts with shared label space. Class-IL, the most challenging setting, requires the model to classify among all the learned classes without task identity. In this work, we focus on Class-IL setting.  

Catastrophic forgetting in CL is typically addressed using memory replay, regularisation based constraints, or dynamic architectural expansion \cite{ebrahimi2020adversarial, aljundi2017expert, li2017learning}. Earlier CL models were trained from scratch, demanding substantial time and computational resources. In contrast, the use of pre-trained models, trained on large scale datasets like ImageNet has emerged as a practical alternative, enabling efficient task adaptation while minimising resource consumption and training effort. These models provide strong general-purpose representations and can be fine-tuned to integrate domain-specific knowledge, thereby enhancing the performance while preserving the prior learning. This shift facilitates scalable and resource efficient continual learning, especially in scenarios requiring rapid deployment. 

In this paper, we present Continual Learning via Text-Image Synergy (CLTS), a novel approach designed to address key limitations in existing CL models, including catastrophic forgetting, prolonged training durations, and high resource and memory demands. CLTS employs architectural expansion to mitigate forgetting while integrating pre-trained models to reduce reliance on large-scaled annotated datasets and minimise computational overhead. The key contributions of this work are:

\noindent 1. A modular architecture that supports dynamic architectural expansion, integrating Task Heads (TH), a Task Router (TR), and pre-trained models to enable continual adaptation in the Class-IL setting. 

\noindent 2. A lightweight memory replay strategy that retains textual captions of prior tasks instead of raw images, substantially reducing the memory requirements without sacrificing the performance.

\noindent 3. A novel use of text-to-image generation: training the Task Router (TR) using synthetic images generated by the Stable Diffusion ($SD$) model, conditioned on stored captions.

We benchmark CLTS on CIFAR10, CIFAR100 and TinyImagenet against four state-of-the-art baselines. Results show that CLTS consistently outperforms them, effectively mitigating catastrophic forgetting.
\section{Related Works}
Continual learning approaches are broadly classified into replay-based, regularization-based, and architectural expansion methods. Replay techniques mitigate forgetting by mixing stored or generated exemplars from previous tasks with new data \cite{rebuffi2017icarl, rolnick2019experience, ebrahimi2020adversarial, rostami2020generative}. Regularization imposes constraints to preserve prior knowledge \cite{li2017learning, kirkpatrick2017overcoming}. Architectural methods grow the network to accommodate new tasks \cite{rusu2016progressive, mallya2018packnet, aljundi2017expert}. Diffusion-based CL methods like SDDR \cite{jodelet2023class} and DDGR \cite{gao2023ddgr} still depend on memory replay and require task labels at inference---limiting scalability and autonomy. 

CaSSLe \cite{fini2022self} introduces an auxiliary predictor for knowledge distillation, aligning current representations with the prior ones. Although compatible with other self-supervised learning (SSL) methods, it relies on known task boundaries at inference, rendering it unsuitable for Class-IL. SCALE \cite{yu2023scale} combines contrastive learning with pseudo-supervision, a self-supervised forgetting loss, and uniform replay sampling. However, its performance remains sensitive to quality of replay data and chosen SSL backbone. U-TELL \cite{solomon2024u} employs task experts, a task assigner, and structured data generators to model task distributions without storing individual samples. While it performs adequately on digit datasets, it fails on real-world data due to generation of low quality images. Self-evolving clustering networks with dynamic architectures have been proposed \cite{pratama2021unsupervised, senthilnath2024evolving} to support online learning via on-the-fly neuron and cluster creation. Centroid-based replay is used to address forgetting in discrete latent spaces with limited generative support, but results remain weak on real-world benchmarks. UPL-STAM \cite{smith2021unsupervised} replaces exemplars with stored centroids, using a pipeline of clustering, novelty detection, and selective forgetting. However, it heavily depends on pre-processing and memory management, limiting robustness and scalability. 

Recognising the constraints of earlier modular approaches in CL \cite{pratama2021unsupervised,aljundi2017expert,solomon2024u,senthilnath2024evolving}, we present CLTS--- a scalable framework that integrates pre-trained modules and avoids the limitations of task-specific replay or data generation. In contrast, our method discards both exemplar storage and test-time labels, using caption conditioned diffusion to train a Task Router, which enables task inference by dynamically routing the inputs to appropriate Task Heads. 

\section{Proposed Architecture}
\begin{figure}[!ht]
\centering
\fontsize{8}{10}\selectfont
\begin{subfigure}[b] {0.48\textwidth}
         \centering         \includegraphics[width=0.95\linewidth]{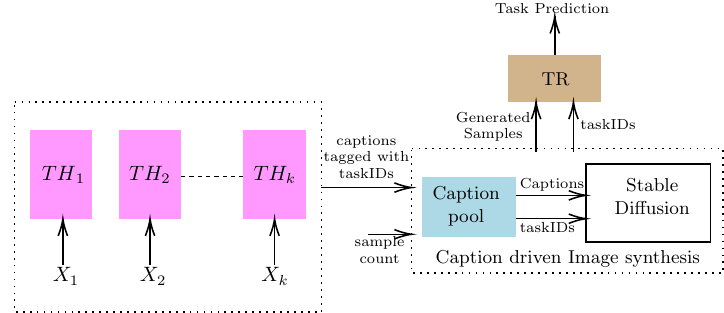}
         \captionsetup{font=footnotesize}
         \caption{}
         \label{fig:blockdia}
\end{subfigure}
\begin{subfigure}[b] {0.38\textwidth}
            \centering
\includegraphics[width=1\linewidth]{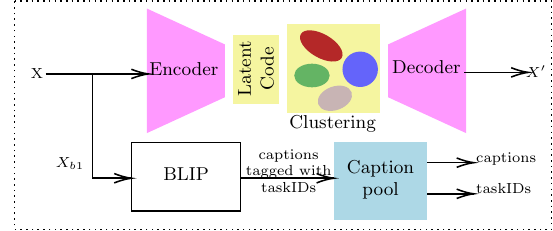} 
         \captionsetup{font=footnotesize}
         \caption{}
         \label{fig:taskhead }
         
\end{subfigure}
\captionsetup{font=footnotesize}
\caption{CLTS architecture: (a) Block diagram; (b) Architectural details of Task Head network}
% (c) Task predictor with SD model.}
\label{fig:fig1}
\Description[Architecture]{Describes 3 blocks of CLTS architecture.}
\end{figure}
\noindent  \textbf{Problem setting.}
% \subsection{Preliminaries}
% We begin with the preliminaries to provide the necessary context for understanding our proposed architecture.
We consider a class-incremental learning (Class-IL) setup, where the task sequence is defined as  $\tau =\{\tau_t\}_{t=1}^k$, with each task $\tau_t$ comprising $n_t$ samples and a disjoint label space. The input data for task $\tau_t$ is denoted as $X_t = {x_j \in \mathbb{R}^{n \times m \times p}}$. A low-dimensional latent representation of task data is denoted as $Z_t \in \mathbb{R}^d$, with $d \ll n \times m \times p$. Additionally, $\mathcal{X}_t$, $t$ refer to synthetic samples generated by Stable Diffusion ($SD$), and their associated taskIDs respectively. The generated captions are denoted by $C_t$, while TH and TR represent the Task Head and Task Router, respectively.

% which is mathematically expressed as,
% \begin{equation}\label{task_il}
% t\neq {t^{\prime}} \Rightarrow Y_t\cap Y_{t^{\prime}} = \emptyset.
% \end{equation}

% \normalsize
% In Domain-IL the class labels for all the tasks are the same, 
% $Y_{t} = Y_{t^{\prime}}.$
% % \small
% % \begin{equation}\label{domain_il}
% % Y_{t} = Y_{t^{\prime}}.
% % \end{equation}
% \normalsize
% % One batch of labeled training data is provided to associate the class labels to the clusters and to extract the task signatures. 
% Task expert and task assigner are denoted by $TE$ and $TA$ respectively. 
% 
\subsection{CLTS Architecture}
The proposed CLTS framework (Fig. \ref{fig:blockdia}) comprises two core components: the Task Head (TH) and the Task Router (TR), alongside a pre-trained Bootstrapping Language-Image Pre-training (BLIP) \cite{li2022blip} and Stable Diffusion ($SD$) \cite{rombach2022high} module. The TH captures task-specific distributions in a low-dimensional latent space and uncovers hidden patterns in the feature embeddings via $K$-Means clustering. Under the Class-IL setting, where task identity is unavailable at test time, the TR predicts the task distribution of incoming samples and routes them to the appropriate TH. An initial mini-batch of labelled data seeds the clustering process, allowing the construction of a lookup table that maps clusters to class labels. This mapping is used during inference for label recovery.

The CLTS training proceeds in two-stages. First, the TH networks are incrementally trained as task data arrives, encoding inputs into a low-dimensional latent space where clustering is performed. In the second stage, the TR is trained using synthetic samples generated by prompting a pre-trained SD model with task-specific captions. These captions are generated using BLIP and each synthetic sample is labelled with its original taskID. Both SD and BLIP are kept frozen, having been pre-trained on large-scale, domain-relevant datasets. The overall training workflow is outlined in Algorithm \ref{alg:CLTS_alg}.

\begin{algorithm}[tb]
% \small
% \SetAlgoLined
% \SetKwInOut{Input}{input}
% \SetKwInOut{Output}{output}
% Input{ Task data $\{X_{t}\}$, $t=1,\ldots,k$}\\
% % \textbf{Parameter}: Optional list of parameters\\
% Output {Task Heads $\{TH_{t}\}$, $t = 1,\ldots,k$ and Task Router $TR$}
\begin{algorithmic} [1] %enables line numbers
\REQUIRE  Task data $\{X_{t}\}$, $t=1,\ldots,k$
\ENSURE Task Heads $\{TH_{t}\}$, $t = 1,\ldots,k$ and Task Router $TR$
\STATE Let $X_{t}$, $t=1, \dots, k$ is the sequential task data
\WHILE{$t$}
\STATE Train $TH_t$ using $X_{t}$ and obtain latent representations $Z_t$
\STATE Generate captions for the initial mini-batch of task training data with pre-trained BLIP captioning model \\$C_{t_{b_1}} = BLIP(X_{t_{b_1}})$
\STATE Incrementally update the caption pool by adding captions tagged  with taskIDs $t$ as $\{C_{t_{b_1},t}\}$.
\STATE Perform clustering on latent data $Z_t$
\ENDWHILE
\FOR{$t \le k$}
\STATE Generate task samples using SD model, $\mathcal{X}_{t} = SD(C_{t})$ and assign taskID $t$, $t=1,\ldots,k$
\STATE Resize $\mathcal{X}_{t}$ to the size of $X_t$
\ENDFOR
\STATE Train Task Router $TR$ with $\{(\mathcal{X}_{t},t)\}$, $t=1,\ldots,k$
% \STATE \textbf{return} solution
\end{algorithmic}
\caption{Continual Learning via Text-Image Synergy (CLTS)}
\label{alg:CLTS_alg}
\end{algorithm}
% \vspace{1mm}
\noindent \textbf{Task Head (TH).} The Task Head (TH) processes a stream of sequential tasks by instantiating a new module upon the arrival of a new task. TH comprises of three key components: i) variational autoencoder (VAE) that learns a low-dimensional latent representation of the task data distribution, enabling efficient encoding; ii) a $K$-Means clustering block that organizes these latent task data into semantically meaningful clusters; and iii) a BLIP module, pre-trained on large-scale datasets, which generates descriptive captions for training samples. Unlike conventional CL models that store task samples for replay, CLTS replaces data storage with caption storage, using these captions to prompt image generation . This significantly reduces memory overload. The TH is trained on each incoming task and generates between 60---100 descriptive captions, depending on the training batch size. Each caption is tagged with a corresponding task identifier (taskID) and stored in a dynamic caption pool. This pool is incrementally updated as new tasks arrive. The stored taskIDs are subsequently used to train the Task Router (TR) in a supervised manner. Fig. \ref{fig:taskhead } illustrates the TH architecture, highlighting its components. 

\noindent{\textbf{Task Router (TR).}}
The Task Router (TR) module is responsible for selecting the most suitable TH to process each incoming test sample. TR is trained in a supervised manner using synthetic samples generated by the Stable Diffusion (SD) along with their associated taskIDs. These taskIDs are distinct from the class labels and are algorithmically assigned based on the task information inferred from the text captions. For the TR, we employ a VGG19 model pre-trained on ImageNet, which is then fine-tuned by replacing its final two layers. The output layer is configured to produce logits corresponding to the total number of tasks.

\noindent{\textbf{Training Objective}
The training objective of CLTS involves minimization of two key loss functions. The total loss is defined as:

\begin{equation}\label{total_loss}
 \mathcal{L} = \lambda_1 \sum_t \mathcal{L}_{TH}^{(t)}+\lambda_2 \mathcal{L}_{TR},
\end{equation}
 where $\lambda_1$ and $\lambda_2$ are tunable hyperparameters. For each task $t$, the loss $\mathcal{L}_{TH}^{(t)}$ comprises of two components: 
\begin{equation}\label{th_loss}
\mathcal{L}_{TH}^{(t)} = \mathcal{L}_{vae}+\mathcal{L}_{clust},  \\
\end{equation}
where $\mathcal{L}_{vae}$ is the encoder-decoder loss of $VAE$ and
the second term $\mathcal{L}_{clust}$ is a clustering loss given by: 
\begin{equation}\label{clust_loss1}
 \mathcal{L}_{clust} = \sum_{j=1}^{J} \sum_{z_i \in M_j} \lVert \mathbf{z_i} - \mathbf{\mu_j}\rVert^2
 \\
 \end{equation}
The loss $\mathcal{L}_{TR}$ trains the TR module to predict taskIDs based on the synthetic inputs. The loss $\mathcal{L}_{TR}$ is defined as a cross-entropy loss between reference taskIDs $t_i$ and the predicted taskIDs $p_i$.  
\begin{equation}\label{tr_loss1}
 \mathcal{L}_{TR} = 
-\sum_{i}t_i\log(p_i) ,  \\
 \end{equation}

% The second term $\mathcal{L}_{TH}$, in eq. \ref{total_loss} is the $TH$ loss function and expressed as,
% \begin{equation}\label{tp_loss1}
%  \mathcal{L}_{TH} = 
% -\sum_{i}c_i\log(p_i) ,  \\
%  \end{equation}
% where $c_i$ is the pseudo label and $p_i$ is the predicted task label. The loss term $\mathcal{L}_{TR}$ is the cross entropy loss between the pseudo label $c_i$ and predicted task label $p_i$.

\noindent{\textbf{Testing Procedure.}}
In the testing phase of the CLTS model, TR, Encoder block of VAE and the $K$-Means clustering module work in unison to process the incoming test samples. The TR first predicts the task, $t$ for each incoming test sample and the corresponding $TH_t$ is selected for processing. The VAE encoder then transforms these test samples into low-dimensional latent representations, which are subsequently grouped into clusters by the K-Means algorithm. Finally, these clusters are cross-verified with the pre-constructed lookup table for validation.

% \begin{figure}[!ht]
% \centering
% \fontsize{8}{10}\selectfont
% \includegraphics[width=0.35\textwidth]{figures/uscl_figure2.pdf} % Reduce the figure size so that it is slightly narrower than the column.
% \captionsetup{font=normalsize}
% \caption{Block diagram describing testing phase of CLTS.}
% \label{Fig2_testingblockdia}
% \Description{Testing process}
% \end{figure}

\section{Experimental Results}
We evaluate CLTS model under Class-IL setting using average task accuracy ($ACC$) as primary metric, $ACC = \frac{1}{k}\sum_{t=1}^k A_{t}$, consistent with prior works \cite{ebrahimi2020adversarial, zhou2024class}. $ACC$ captures the model's performance across all incremental stages, with higher values indicating stronger performance across multiple tasks throughout these stages \cite{zhou2024class}.
% $ ACC = \frac{1}{k}\sum_{t=1}^k A_{t}$
% \begin{equation}\label{avg_acc}
%  ACC = \frac{1}{k}\sum_{t=1}^k A_{t},  \\
% \end{equation}
 % where $k$ is the total number of tasks and $A_t$ is the test accuracy for task $t$.

\noindent{\textbf{Datasets.}}
We evaluate the performance of the CLTS model across three real-world Class-IL benchmarks: Split CIFAR10 (SCIFAR10) \cite{pratama2021unsupervised}, Split CIFAR100 (SCIFAR100) and Split TinyImageNet (STinyImageNet) \cite{yu2023scale}. Among the chosen benchmarks, the SCIFAR100 datastream provides a larger number of tasks, enabling a detailed analysis of our model's performance. The SCIFAR100 datastream is structured as a sequence of 10 tasks ($t_1,\ldots,t_{10}$), with each task introducing two new classes out of a total of 20 coarse classes \cite{yu2023scale}. Within Class-IL setting, we organize the SCIFAR10 \cite{pratama2021unsupervised} and STinyImageNet datasets into a sequence of five tasks $(t_1, t_2, t_3, t_4, t_5)$. Each task is defined by a unique pair of mutually exclusive classes $(\{\{0,1\}, \{2,3\},\{4,5\}, \{6,7\}, \{8,9\}\})$.

\begin{table}[!h]
\centering
\fontsize{7}{9}\selectfont
\setlength{\tabcolsep}{3pt}
\captionsetup{font=footnotesize}
\caption{Performance comparison of CLTS with selected baselines on CIFAR10, CIFAR100 and TinyImagenet datasets}
\begin{tabular}{cccc}
\hline
\multirow{2}{*}{\textbf{Technique}} & \multicolumn{3}{c}{\textbf{Dataset}}  \\ \cmidrule{2-4} & \multicolumn{1}{c}{SCIFAR10} & \multicolumn{1}{c}{SCIFAR100}  
&\multicolumn{1}{c}{STinyImageNet}  \\
\hline
SCALE &21.72$\pm$1.01  &13.68$\pm$0.78   &21.66$\pm$0.68  \\
CaSSLe &16.91$\pm$1.00  &10.97$\pm$0.84 &20.66$\pm$1.35 \\
UPL-STAM &28.02$\pm$1.99  &13.22$\pm$0.32    &21.68$\pm$1.30  \\
U-TELL &29.65$\pm$0.67 &16.79$\pm$0.38
&27.78$\pm$1.13  \\
\textbf{CLTS(ours)} &\textbf{32.27$\pm$1.18} &\textbf{25.86$\pm$0.80}    &\textbf{35.6$\pm$1.22}  \\
\hline
\end{tabular}
\label{tab:perf_cmpr} 
\end{table}

\noindent{\textbf{Baselines.}}
We chose U-TELL \cite{solomon2024u}, SCALE \cite{yu2023scale}, CaSSLe \cite{fini2022self} and UPL-STAM \cite{smith2021unsupervised} as baselines for our experimental study.  
The CaSSLe model originally designed under the assumption of known task boundaries, was modified to function within unknown task boundaries to ensure compatibility with the Class-IL framework.

\noindent{\textbf{Performance Comparison.}}
Table \ref{tab:perf_cmpr} compares the CLTS model against selected baselines.  The reported values are average task accuracy computed from the mean and standard deviation over ten runs. The CLTS model outperforms the best performing baseline by a considerable margin. The Class-IL experiments on SCIFAR10 show an $8.8\%$ improvement in average task accuracy. The SCIFAR100 dataset (10 tasks) demonstrates an improvement of $54.02\%$, while STinyImageNet dataset shows a $28.15\%$ improvement. These results are driven by integrating the pre-trained models into the architecture. The SD module generates a diverse, high quality image dataset that trains the TR module, thereby improving the task prediction and TH selection. Fig. \ref{fig:genimages_cifar10} and \ref{fig:genimages_tiny} show a few examples of BLIP-generated captions, along with corresponding diverse images generated by SD for the SCIFAR10 and STinyImageNet datasets, respectively.
\begin{figure}[!ht]
\centering
\fontsize{8}{10}\selectfont
\begin{subfigure}[b] {0.23\textwidth}
         \centering         \includegraphics[width=0.95\linewidth]{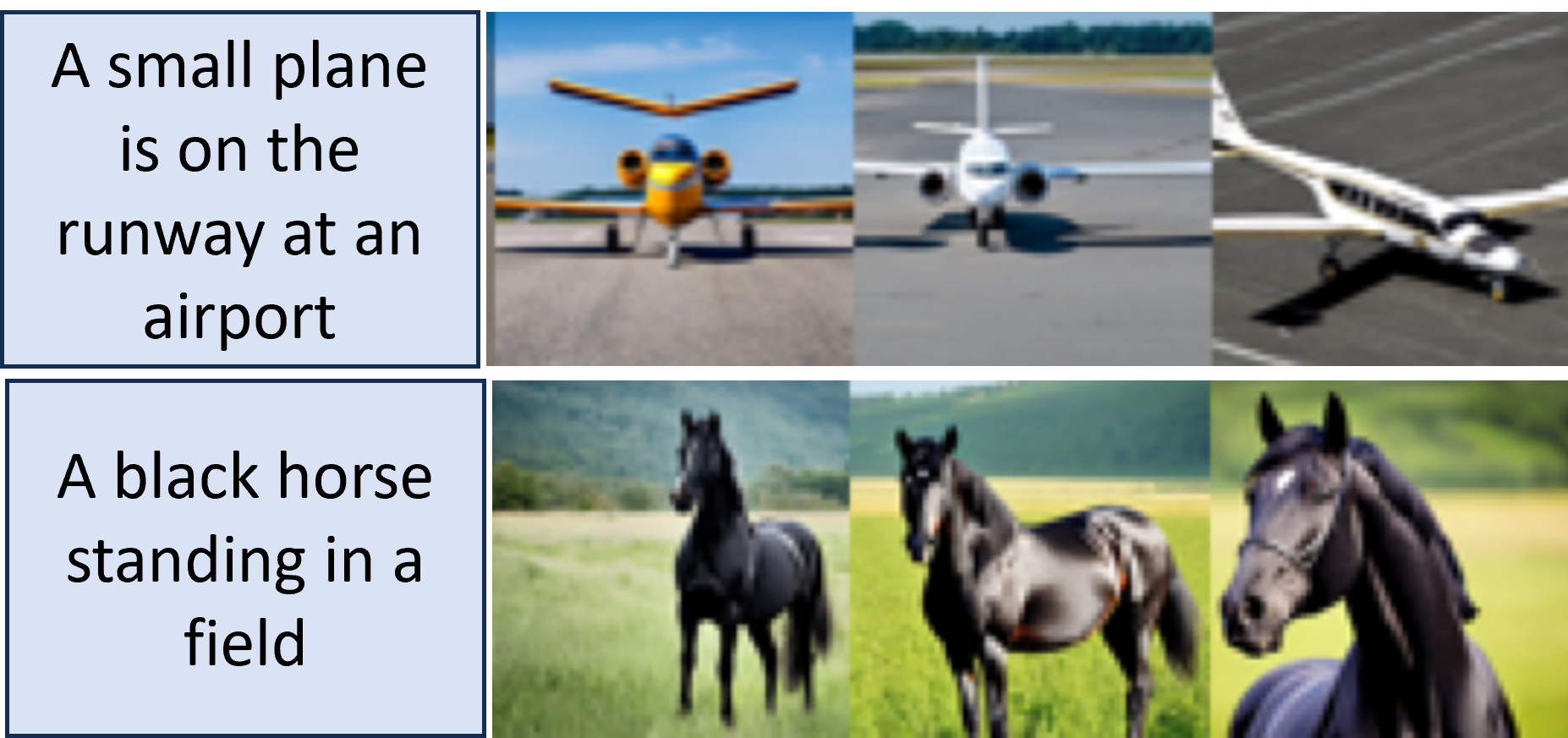}
         \captionsetup{font=footnotesize}
         \caption{}
         \label{fig:genimages_cifar10}
\end{subfigure}
\begin{subfigure}[b] {0.23\textwidth}
            \centering
\includegraphics[width=1\linewidth]{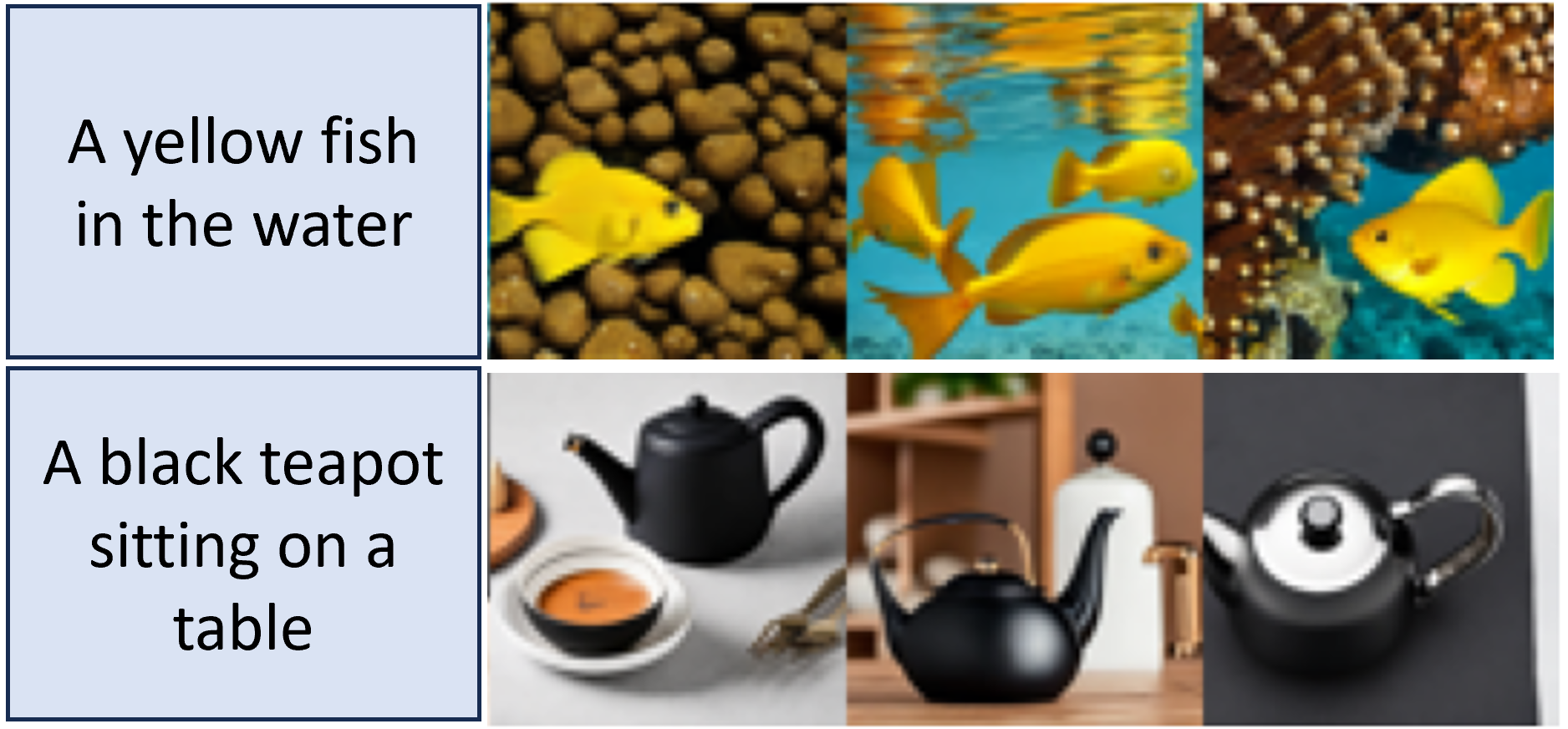} 
         \captionsetup{font=footnotesize}
         \caption{}
         \label{fig:genimages_tiny}
         
\end{subfigure}
\captionsetup{font=footnotesize}
\caption{Examples of generated images from text prompt: (a) SCIFAR10; (b) STinyImageNet }
% (c) Task predictor with SD model.}
\label{fig:genimageswithcaptions}
\Description[Generation quality]{Generated images from captions.}
\end{figure}
% \vspace{-0.2 cm}
% Fig. \ref{fig:blipcations} illustrates a selection of example captions generated by the BLIP model for the SCIFAR10 dataset, highlighting its ability to interpret and describe complex images. In Fig. \ref{fig:sd_gen_images}, we see the images generated by the SD model from the text captions. 
\begin{figure}[!ht]
\centering
\fontsize{8}{10}\selectfont
\includegraphics[width=0.43\textwidth]{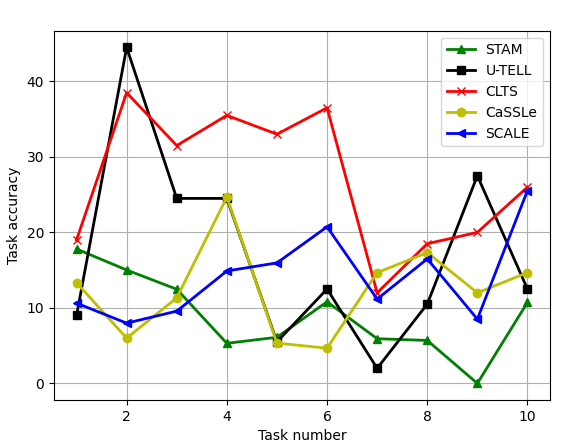} % Reduce the figure size so that it is slightly narrower than the column.
\captionsetup{font=footnotesize}
\caption{Individual task performance comparison of CLTS and baselines on SCIFAR100 dataset.} 
\label{Fig3_taskplot}
\Description{Task performance comparison for SCIFAR100}
\end{figure}

Fig. \ref{Fig3_taskplot} compares the task-wise performance of the CLTS model and baselines on SCIFAR100 dataset under expanding clustering tasks. Despite SCIFAR100's increased task count, the CLTS model maintains stable performance, showing minimal catastrophic forgetting and outperforming the baselines on 7 out of 10 tasks. Table \ref{tab:mem_eff} compares the memory efficiency of the CLTS model against the selected baselines, based on the memory required to store previous task samples or centroids for replay. Unlike the baselines, which rely on storing the task exemplars or centroids, the CLTS model maintains only a pool of text captions, resulting in negligible memory usage. 

We report the key hyperparameters used in our experiments. All models use a batch size between 60---100, learning rate of 0.0002, and Adam optimizer. The loss weights $\lambda_1$ and $\lambda_2$ are set to 1.0, the number of clusters varies from 2---20, and the latent dimension is fixed at 128. Training epochs range between 50---350 depending on the dataset. The TH module has 3 layer CNN architecture, while the BLIP and SD modules are used as frozen without fine-tuning. 
\begin{table}[!h]
\centering
\fontsize{7}{9}\selectfont
\setlength{\tabcolsep}{3pt}
\captionsetup{font=footnotesize}
\caption{Memory efficiency comparison in MB}
\begin{tabular}{ccccc}
\hline
\multirow{2}{*}{} & \multicolumn{4}{c}{\textbf{Method}}  \\ \cmidrule{2-5} \multicolumn{1}{c}{Dataset} & \multicolumn{1}{c}{SCALE} & \multicolumn{1}{c}{UPL-STAM}  
& \multicolumn{1}{c}{U-TELL} &\multicolumn{1}{c}{\textbf{CLTS}} \\
\hline
SCIFAR10 &15.72 \textcolor{red}{(+6287$\%)$} &3.09 \textcolor{red}{(+1235$\%$)}     &{0.17} \textcolor{red}{(+67$\%$)}  & \textbf{0.0025}\\
STinyImageNet &62.91 \textcolor{red}{(+1497$\%$)} &5.36 \textcolor{red}{(+126.62$\%$)} &{0.17} \textcolor{red}{(+3.05$\%$)} &\textbf{0.042}\\
\hline
\end{tabular}
\label{tab:mem_eff} 
\end{table}
% The CLTS experiments are conducted with the number of clusters ranging from 2 to 20. discussions subsection

\noindent \textbf{Ablation Study.}
We conduct an ablation study on CLTS model to analyse the effect of replacing BLIP captioning with two alternatives---ViT+GPT2 and class label names as captions. We also evaluate the effect of replacing SD module with DALL$\cdot$E mini for image generation. ViT+GPT2 model combines a pre-trained ViT (Vision Transformer) image encoder with a GPT2 (Generative Pre-trained Transformer)  decoder, and is primarily fine-tuned for image captioning task \cite{nlp_connect_2022}. In our ablation experiments conducted on SCIFAR10 and STinyImageNet, we used the ViT+GPT2 model as a one-to-one replacement for BLIP. The superior caption quality of BLIP is enabled by its contrastive pre-training and multi-task fine-tuning, led to improved performance of CLTS, as shown in Table \ref{tab:ablation}. In our second experiment, BLIP-generated captions were replaced with single-word class label names. These label-based captions lack the descriptive qualities that SD module relies on, which led to inferior performance as shown in Table \ref{tab:ablation}. In our third experiment, we used DALL$\cdot$Emini as a one-to-one replacement for the SD module to analyse the impact of image generation on CLTS performance. DALL$\cdot$Emini combines a BART-like (Bidirectional and Auto-Regressive Transformer) model for mapping text to image tokens with a pre-trained VQGAN decoder (Vector-Quantized Generative Adversarial Network) for image synthesis \cite{Dayma_DALLE_Mini_2021}. As shown in the Table \ref{tab:ablation}, CLTS performs better with the SD module, due to the higher-quality images generated by the diffusion process, compared to noisy and unnaturally textured images created by DALL$\cdot$Emini. 
\begin{table}[!h]
\centering
\fontsize{7}{9}\selectfont
\setlength{\tabcolsep}{3pt}
\captionsetup{font=footnotesize}
\caption{Ablation study}
\begin{tabular}{ccccc}
\hline
\multirow{2}{*}{} & \multicolumn{4}{c}{\textbf{Method}}  \\ \cmidrule{2-5} \multicolumn{1}{c}{Dataset} & \multicolumn{1}{c}{\makecell{CLTS-BLIP\\+(ViT+GPT2)}} & \multicolumn{1}{c}{\makecell{CLTS-BLIP\\+Label names}}  
& \multicolumn{1}{c}{\makecell{CLTS–SD\\+DALL$\cdot$E mini}} &\multicolumn{1}{c}{\textbf{CLTS}} \\
\hline
SCIFAR10 &26.99 $\pm$ 1.09 &24.68 $\pm$ 2.46  &17.87 $\pm$ 1.12 & \textbf{32.77 $\pm$ 0.97}\\
STinyImageNet &29.55 $\pm$ 1.23 &26.35 $\pm$ 1.30 &-- &\textbf{35.2 $\pm$ 1.43}\\
\hline
\end{tabular}
\label{tab:ablation} 
\end{table}

\noindent \textbf{Discussion.}
The performance of the proposed CLTS model highlights the importance of high-quality, caption conditioned image generation in continual learning. CLTS can handle frequent task arrivals in real-world scenarios by storing only the lightweight task-specific captions and enabling fast adaptation through generative replay, thereby avoiding the need for large memory buffers. On the downside, since the pre-trained models are typically trained on common everyday images (e.g., animals, vehicles, everyday objects), applying CLTS to specialized domains such as satellite imagery, medical imaging, or industrial defect detection may require additional fine-tuning of the pre-trained modules, leading to increased training complexity.

\section{Conclusion}
We propose CLTS, a modular continual learning framework that combines Task Heads (TH), Task Router (TR) and pre-trained vision-language modules BLIP and Stable Diffusion (SD). BLIP generates text captions from task training samples, which are stored in caption pool in place of raw image samples. During replay, these captions serve as prompts for SD module to generate synthetic images, enabling TR training. 
In Class-IL setting, CLTS outperforms baselines U-TELL, UPL-STAM, CaSSLe and SCALE on SCIFAR10, SCIFAR100 and STinyImageNet datasets. CLTS architecture supports scalable task expansion, and rapid adaptation, making it suitable for evolving real-world scenarios.   
\section{GenAI Usage Disclosure}
ChatGPT (OpenAI, GPT-4o, 2025) was used to clarify language, composition, and grammar corrections, under the supervision and final editorial control of the authors. No AI-generated content was used without thorough human review, and the tool was not used to generate code, figures, or fabricate data, results.
% \section{Acknowledgments}

% Identification of funding sources and other support, and thanks to
% individuals and groups that assisted in the research and the
% preparation of the work should be included in an acknowledgment
% section, which is placed just before the reference section in your
% document.
\bibliographystyle{ACM-Reference-Format}
\bibliography{sample-base}

% %%
% %% If your work has an appendix, this is the place to put it.
% \appendix









\end{document}